\documentclass[conference]{IEEEtran}
\IEEEoverridecommandlockouts

\usepackage{cite}
\usepackage{amsmath,amssymb,amsfonts}
\usepackage{algorithmic}
\usepackage{graphicx}
\usepackage{textcomp}
\usepackage{xcolor}
\usepackage{placeins}
\usepackage{slashbox}
\usepackage{times}
\usepackage[utf8]{inputenc}
\usepackage[main=english, vietnamese]{babel}
\babeltags{vi=vietnamese}
\usepackage{type1cm}
\usepackage[T1]{fontenc}
\usepackage{microtype}
\usepackage{multirow}
\usepackage{graphicx}
\usepackage{marvosym}
\usepackage{float}
\usepackage{amsmath,amssymb,amsfonts}
\usepackage{pifont}
\usepackage{siunitx}
\usepackage{latexsym}
\usepackage{etoolbox}

\usepackage[bookmarks,bookmarksopen,bookmarksdepth=3,bookmarksnumbered]{hyperref}
\hypersetup{colorlinks, citecolor=blue, linkcolor=blue, urlcolor=blue}

\begin{document}

\title{Leveraging Semantic Representations Combined with Contextual Word Representations for\\Recognizing Textual Entailment in Vietnamese}

\author{\IEEEauthorblockN{\textbf{Quoc-Loc Duong$^{\text{1, 2}}$, Duc-Vu Nguyen$^{\text{1, 2}}$, Ngan Luu-Thuy Nguyen$^{\text{*, 1, 2}}$}}
\IEEEauthorblockA{$^{\text{1}}$University of Information Technology, Ho Chi Minh City, Vietnam \\
$^{\text{2}}$Vietnam National University, Ho Chi Minh City, Vietnam\\
\texttt{18521006@gm.uit.edu.vn}\qquad \texttt{\{vund,ngannlt\}@uit.edu.vn}}
\thanks{$^{\text{*}}$Corresponding author}
}

\maketitle

\begin{abstract}
RTE is a significant problem and is a reasonably active research community. The proposed research works on the approach to this problem are pretty diverse with many different directions. For Vietnamese, the RTE problem is moderately new, but this problem plays a vital role in natural language understanding systems. Currently, methods to solve this problem based on contextual word representation learning models have given outstanding results. However, Vietnamese is a semantically rich language. Therefore, in this paper, we want to present an experiment combining semantic word representation through the SRL task with context representation of BERT relative models for the RTE problem. The experimental results give conclusions about the influence and role of semantic representation on Vietnamese in understanding natural language. The experimental results show that the semantic-aware contextual representation model has about 1\% higher performance than the model that does not incorporate semantic representation. In addition, the effects on the data domain in Vietnamese are also higher than those in English. This result also shows the positive influence of SRL on RTE problem in Vietnamese.


\end{abstract}


\section{Introduction}
According to RTE-1 conference, the Recognizing Textual Entailment (RTE) task is defined as recognizing, given two text fragments, whether the meaning of one text can be inferred (entailed) from the other \cite{RTE1}. This problem plays an essential role in the field of Natural Language Understanding (NLU). Vietnamese has many meanings, leading to ambiguity in spoken or written text. However, this ambiguity is little noticed in practice because humans can handle this phenomenon well. Ambiguity is the phenomenon of ambiguity of ideas that blur the line between this and that. The ambiguity occurs mainly in the Vietnamese language and is inevitable as a general rule. Many NLU problems such as Question Answering, Text Summarization, or Information Extraction will face difficulties in processing when encountering ambiguity. Therefore, the RTE will help to improve the reasoning problems when facing this phenomenon. In addition, it is used as a metric to evaluate the understanding of the language of the NLP systems in general.

On the other hand, as mentioned, Vietnamese is a semantically rich language that can cause ambiguity. Therefore, we decided to experiment with leveraging semantic representation via the Semantic Role Labeling (SRL) task to combine with contextual representation via transformer-based encoder models. This experiment can be considered the latest approach to the RTE problem in Vietnamese. On the other hand, taking advantage of context representation is the primary approach to the RTE problem in Vietnamese, and has yielded outstanding results.

\begin{figure}[]
\centering
\includegraphics[scale=0.7]{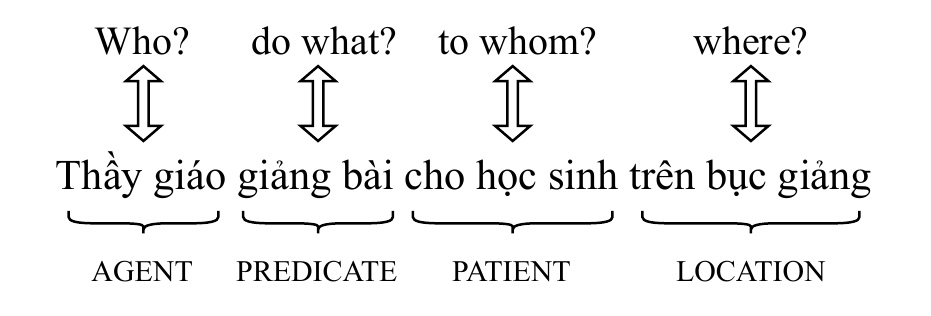}
\caption{SRL of Vietnamese sentence.}
\label{fig:SR}
\end{figure}

\begin{figure*}[!ht]
\centering
\includegraphics[scale=1.0]{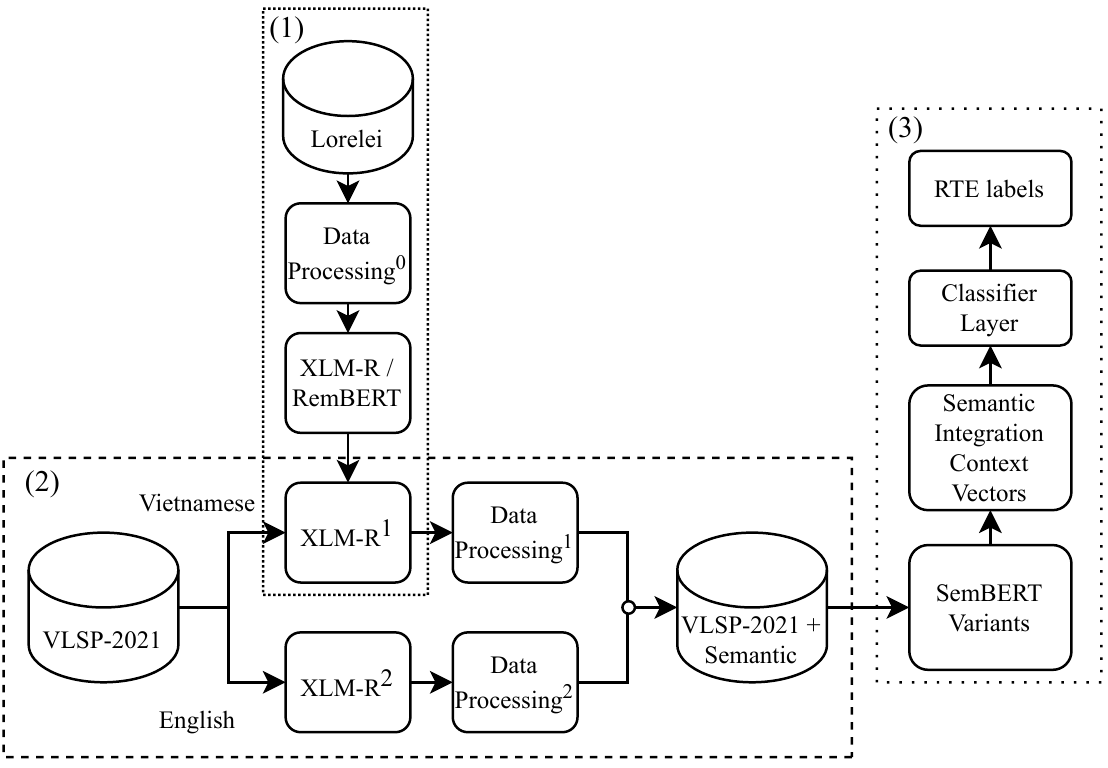}
\caption[Minh họa quá trình triển khai thực hiện tinh chỉnh mô hình biến thể SemBERT cho bài toán Gán nhãn vai nghĩa.]{\label{fig:QuaTrinhThucHienFull}Illustrate the implementation process. (1) The process of fine-tuning the XLM-R model for the SRL task. (2) Semantic information extraction process for the VLSP-2021 corpus. (3) The process of fine-tuning the SemBERT variant model for the RTE problem.}
\end{figure*}

In some theories of linguistics, Semantic Roles (SR) are understood as the semantic roles of nouns, noun phrases, or other components of sentences, collectively known as arguments, for action or state, which is described by a verb or called a predicate. SRL is the problem of determining the meaning roles for sentence components to help clarify the structure and relationship between the predicate and the arguments constituting the sentence. In essence, SRL is a transition problem but plays a crucial role in NLP system. For systems, to efficiently use the information encoded in the text, they must be able to detect the events being described. Therefore, semantic roles help to represent general semantics on the surface of a sentence structure that allows for making simple inferences that cannot be inferred by parsing trees or other methods. Especially, SRL helps the NLU problems improve understanding of natural language based on the structural surface by answering the question of \textit{who did what to whom}, \textit{when}, or \textit{why}. On the other hand, as mentioned, Vietnamese is a semantically rich language; capturing the common semantic representation can improve model performance for NLU problems, namely RTE.

The process of performing our experiment consists of three stages (Figure \ref{fig:QuaTrinhThucHienFull}): 1) Fine-tune the model for the SRL problem; 2) Extract semantic labels for the corpus using the model fine-tuned in the first stage; 3) Fine-tuning the semantic cognitive context representation model for the RTE problem. Through this article, we have approached the RTE problem with a new approach to Vietnamese. In addition, the role of SRL through experimental results can create a premise for developing this problem in Vietnamese.

\section{Related Works}

The RTE problem has attracted the attention of the research community in the field of NLP around the world; notably the RTE-1 \cite{RTE1} - RTE-7 \cite{RTE7} conference series held continuously from 2005 to 2011. The series of conferences aimed to propose methods helping to improve results and develop data for this task. However, the RTE problem is relatively new in Vietnamese but also attracts the research community. In 2021, the VLSP conference \cite{nli2021-vlsp} introduced the first dataset for the RTE. The challenge for this dataset is to include both Vietnamese and English. However, there is a limitation on the approach to the RTE problem at the VLSP-2021 conference.


Since Vaswani et al introduced the Transformer architecture in 2017 \cite{vaswani2017attention}, a series of models have leveraged the coding component of this architecture for contextual word representation learning (BERT \cite{devlin2018bert}, XLM-R \cite{conneau2019unsupervised}, and XLNet \cite{yang2019xlnet}), giving outstanding results for the RTE problem compared to traditional deep learning methods. Similarly, some outstanding works such as PhoBERT \cite{phobert} and BERTweet \cite{bertweet} have also brought good results for the Vietnamese domain. Although these models give good results, they do not consider the semantic side of the input text.


On the other hand, there are hardly any works on exploiting semantic representations through the SRL problem for NLU problems, specifically RTE in Vietnamese. In contrast, for foreign works, Zhang et al proposed the SemBERT model \cite{zhang2020semantics}, which takes advantage of the context representation of the BERT model combined with the semantic representation to give higher results than the BERT model is about 1\% for NLU problems in English and Chinese. In addition, several works of Shi et al \cite{shi2016knowledge}, and Yih et al \cite{yih2016value} also applied the SRL problem, and achieved promising results.


\section{Methodology}

\subsection{Semantics-aware BERT}
\label{sec:semBERT}
The Semantics-aware BERT or SemBERT model \cite{zhang2020semantics} consists of two components: encoding, and integration, with the encoding component consisting of two modules. The first module consists of a BERT model responsible for learning the contextual representation of the input text. Because the BERT model uses the WordPiece word tokenization method, the input text will be split into word and even subword levels. Therefore, the output of the BERT model will go through the convolutional neural network (CNN) layer to perform the alignment to the word level. Meanwhile, the second module includes the Bi-GRU model \cite{cho2014properties}, which will learn semantic representation through semantic label sequences corresponding to the input text. For the integrated component, it will be responsible for integrating two vectors representing context and semantic representation together to form a vector representing semantic-aware context for fine-tuning downstream tasks.



\begin{figure*}[ht]
\centering
\includegraphics[scale=0.75]{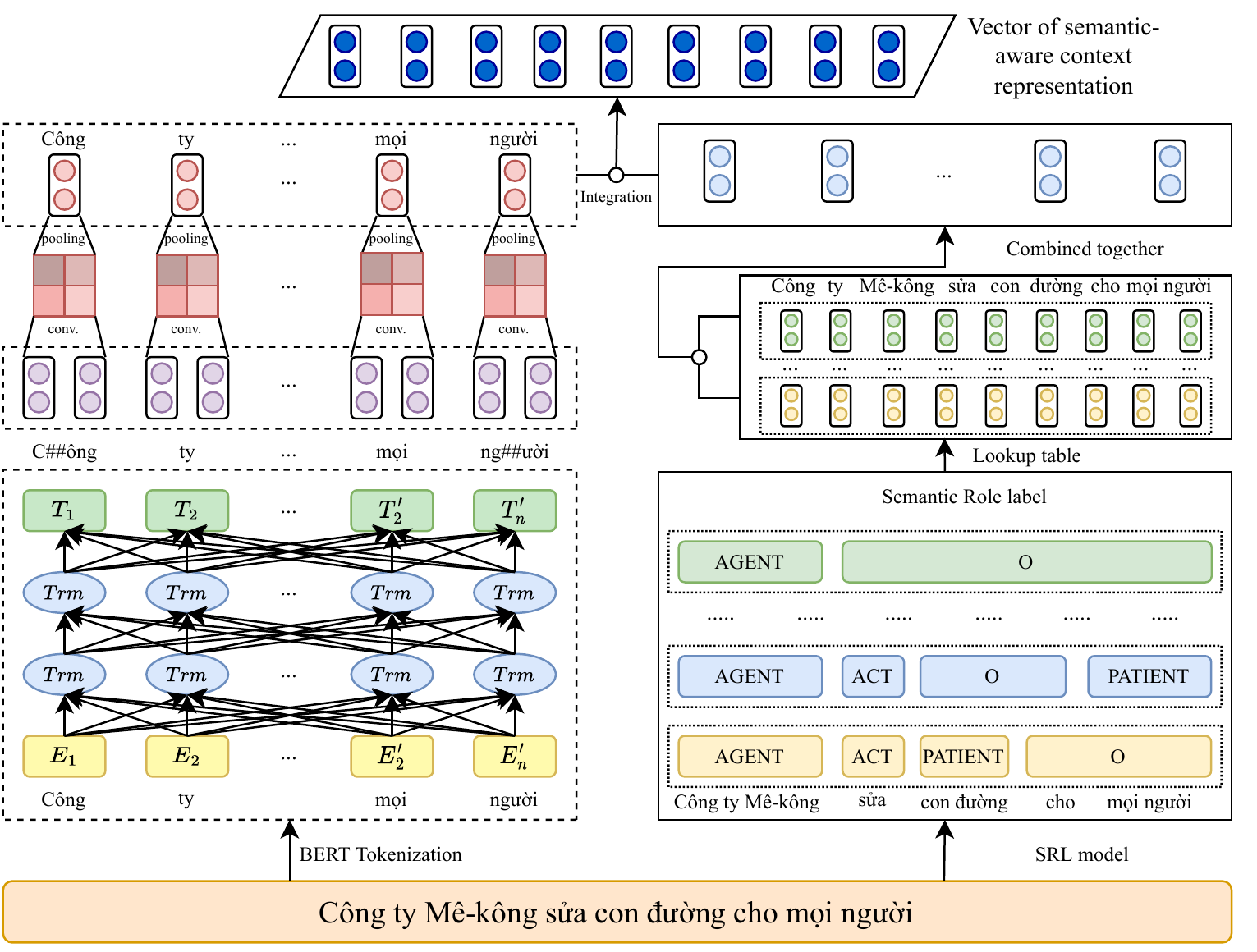}
\caption{Overview of the SemBERT model, the \textit{SRL model} is specifically trained for the problem of SRL with the purpose of extracting semantic labels for text input.}
\label{fig:SemBERT_full}
\end{figure*}

\subsection{SemBERT variants}
We modified the contextual learning module using the BERT model of the SemBERT architecture to the mBERT and XLM-R models. The goal of this modification is because the corpus chosen for the experiment includes Vietnamese and English. On the other hand, both mBERT and XLM-R models have been pre-trained on a large multilingual corpus containing both Vietnamese and English, so learning the context for the input corpus will give better results than the BERT model.

\begin{figure}[!ht]
\centering
\includegraphics[scale=0.6]{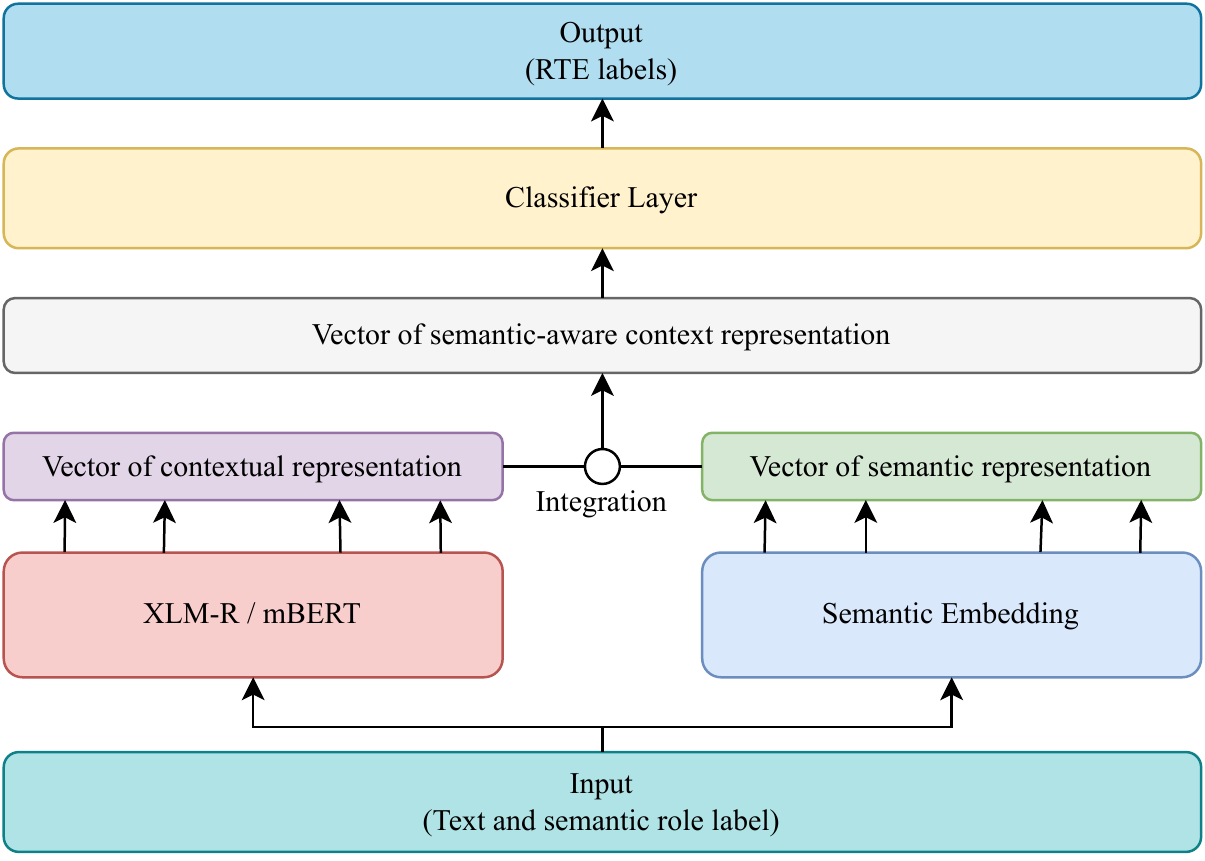}
\caption{Visualization of the SemBERT variant model and the fine-tuning process of the SemBERT variant model.}
\label{fig:FineTuneSemBERT}
\end{figure}

For English, some works have proposed models for extracting semantic labels of texts. However, for Vietnamese, we have difficulty accessing the works on the SRL task to perform this task. Therefore, to extract semantic role labels for input text, we have fine-tuned XLM-R and RemBERT models for the SRL problem. This process aims for the SemBERT variant models to learn the semantic representation described in Section \ref{sec:semBERT}. This task is solved based on the approach of the Sequence Labeling problem with the IOB format of the input (He et al \cite{he2017deep}, and Fei et el \cite{fei2020cross}).



\begin{figure}[!ht]
\centering
\includegraphics[scale=0.45]{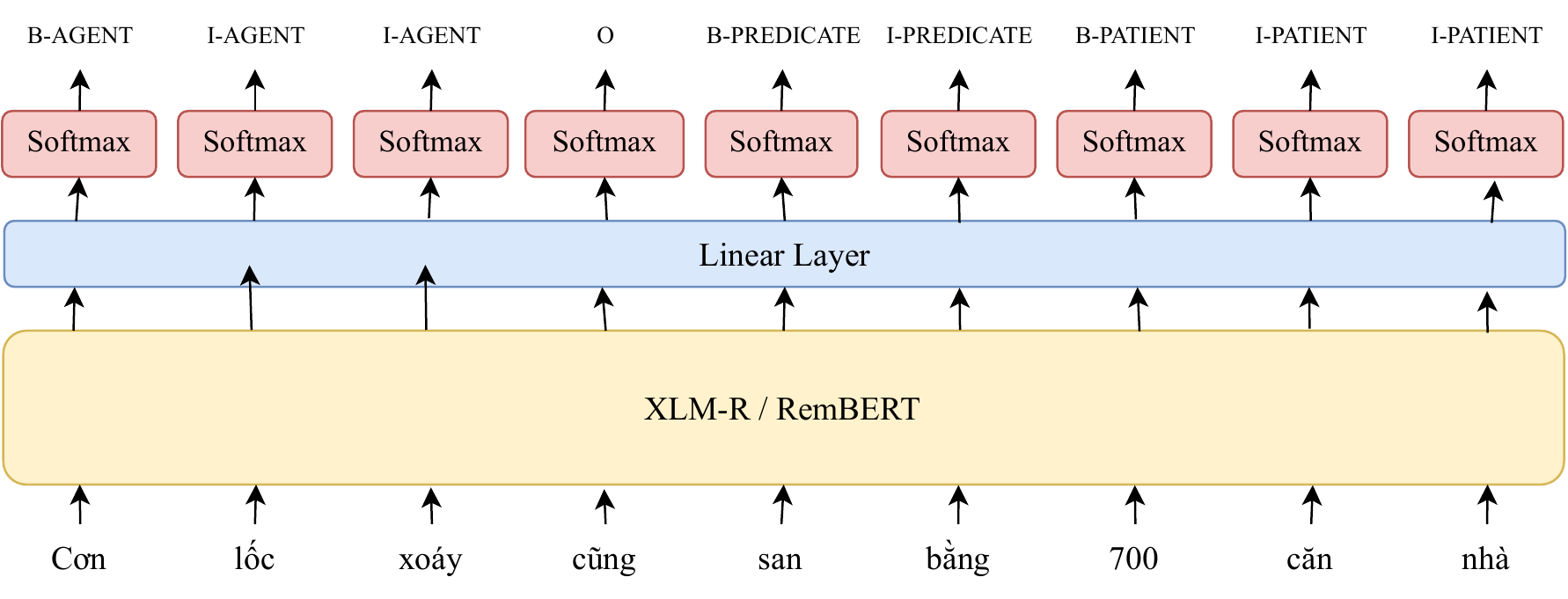}
\caption{Illustrate the process of fine-tuning the XLM-R / RemBERT model for the SRL task.}
\label{fig:IOB_RemBERT}
\end{figure}

\section{Experiments}
The training is performed on the Google Colaboratory virtual machine. In this section, the paper will present the implementation process consisting of three main tasks: Fine-tuning the model for the SRL problem; Extracting semantic labels for input text; Fine-tuning variant models of SemBERT for the RTE problem.


\subsection{Dataset}

\subsubsection{SRL task}
We use the corpus of the LORELEI Language Packs \cite{strassel2016lorelei} for the SRL problem. The dataset for this problem consists of 1760 data points formatted in XML files. According to the author, building this corpus aims to capture a basic understanding of what is happening or the case in a sentence.
\subsubsection{RTE task}
For the RTE problem, we use the VLSP-2021 corpus. The dataset consists of 16,185 training data points and 4177 test data points. Both datasets have an equal distribution of labels and contain Vietnamese and English (Table \ref{table:vi-en-VLSP2021}).

\begin{table}[ht]
\centering
\caption{Statistics Of Vietnamese And English Data Points On The Training And Test Dataset VLSP-2021.}
\label{table:vi-en-VLSP2021}
\resizebox{7.5cm}{!}{%
\begin{tabular}{|c||cc||cc|}
\hline
\multirow{2}{*}{} & \multicolumn{2}{c||}{Training set}          & \multicolumn{2}{c|}{Test set}              \\ \cline{2-5} 
                  & \multicolumn{1}{c|}{\textit{Text\_1}} & \textit{Text\_2} & \multicolumn{1}{c|}{\textit{Text\_1}} & \textit{Text\_2} \\ \hline \hline
Vie                & \multicolumn{1}{c|}{8685}        & 16185       & \multicolumn{1}{c|}{2118}        & 4177        \\ \hline
Eng                & \multicolumn{1}{c|}{7500}        & 0           & \multicolumn{1}{c|}{2059}        & 0           \\ \hline
\end{tabular}%
}
\end{table}

\subsection{Fine-tune the model for the SRL}
We perform a fine-tuning experiment on RemBERT and XLM-R models for the SRL task. The RemBERT model was introduced by Chung in 2020 \cite{chung2020rethinking}, being an improved version of the XLM-R model. In this work, Chung focuses on studying the effect when changing the size of the model's layers. Specifically, they observed that increasing the output size of embedded layers improved performance for fine-tuning the following tasks. In addition, the model is also trained on a large dataset with multiple languages (110 languages). The RemBERT model has achieved about 5\% higher results than the XLM-R model (F1-score) when fine-tuning the Entity Recognition task (this result is directly referenced from the work of the author Chung \cite{chung2020rethinking}).


In natural language, a sentence can have more than one verb (predicate). Thus, arguments can take on different semantic roles for each verb present in a sentence. Therefore, we use the k-fold cross validation method, with \textit{k = 10}, to train ten models for extracting multiple aspects of semantic information from a sentence. The hyperparameters for fine-tuning are set the same for all models: learning rate \textit{1e-5}, batch size \textit{4}, weight decay \textit{0.01}, and epoch number \textit{40} (since the \textit{30\textsuperscript{th}} epoch, there has been no improvement in performance capacity).


Table \ref{table:TinhChinhRemBET-XLMR} shows that the performance of the XLM-R model when fine-tuning for the problem of SRL on LORELEI corpus is higher than that of the RemBERT model. Therefore, we chose the fine-tuned XLM-R model for semantic extraction on the VLSP-2021 corpus.


\begin{table*}[ht]
\centering
\caption{The Results Of The Problem Of SRL After Fine-Tuning Two Models, XLM-R And RemBERT, On The LORELEI Dataset With The k-fold Cross-Evaluation Method ($k = 10$) ($\%$).}
\label{table:TinhChinhRemBET-XLMR}
\resizebox{11cm}{!}{%
\begin{tabular}{|c||ccc||ccc|}
\hline
\multirow{2}{*}{} & \multicolumn{3}{c||}{XLM-R}                                              & \multicolumn{3}{c|}{RemBERT}                                            \\ \cline{2-7} 
                  & \multicolumn{1}{c}{Precision} & \multicolumn{1}{c}{Recall} & F\textsubscript{1}-score & \multicolumn{1}{c}{Precision} & \multicolumn{1}{c}{Recall} & F\textsubscript{1}-score \\ \hline \hline
Model 1         & \multicolumn{1}{c}{36.72}     & \multicolumn{1}{c}{33.62}  & 35.10    & \multicolumn{1}{c}{38.44}     & \multicolumn{1}{c}{32.16}  & 35.02    \\ \hline
Model 2         & \multicolumn{1}{c}{44.83}     & \multicolumn{1}{c}{30.37}  & 36.21    & \multicolumn{1}{c}{37.53}     & \multicolumn{1}{c}{30.19}  & 33.46    \\ \hline
Model 3         & \multicolumn{1}{c}{39.62}     & \multicolumn{1}{c}{31.92}  & 35.36    & \multicolumn{1}{c}{46.00}     & \multicolumn{1}{c}{28.94}  & 35.53    \\ \hline
Model 4         & \multicolumn{1}{c}{41.34}     & \multicolumn{1}{c}{30.21}  & 34.91    & \multicolumn{1}{c}{47.67}     & \multicolumn{1}{c}{29.49}  & 36.43    \\ \hline
Model 5         & \multicolumn{1}{c}{38.82}     & \multicolumn{1}{c}{31.72}  & 34.92    & \multicolumn{1}{c}{35.74}     & \multicolumn{1}{c}{27.62}  & 31.16    \\ \hline
Model 6         & \multicolumn{1}{c}{38.20}     & \multicolumn{1}{c}{32.34}  & 35.03    & \multicolumn{1}{c}{37.56}     & \multicolumn{1}{c}{28.87}  & 32.64    \\ \hline
Model 7         & \multicolumn{1}{c}{36.72}     & \multicolumn{1}{c}{32.20}  & 34.30    & \multicolumn{1}{c}{38.58}     & \multicolumn{1}{c}{30.01}  & 33.78    \\ \hline
Model 8         & \multicolumn{1}{c}{41.28}     & \multicolumn{1}{c}{31.42}  & 35.69    & \multicolumn{1}{c}{37.06}     & \multicolumn{1}{c}{31.15}  & 34.01    \\ \hline
Model 9         & \multicolumn{1}{c}{42.32}     & \multicolumn{1}{c}{33.21}  & 37.22    & \multicolumn{1}{c}{40.47}     & \multicolumn{1}{c}{30.79}  & 34.97    \\ \hline
Model 10        & \multicolumn{1}{c}{40.91}     & \multicolumn{1}{c}{33.67}  & 36.94    & \multicolumn{1}{c}{38.26}     & \multicolumn{1}{c}{33.04}  & 35.46    \\ \hline
\textbf{Average}        & \multicolumn{1}{c}{\textbf{40.08}}     & \multicolumn{1}{c}{\textbf{32.07}}  & \textbf{35.57}    & \multicolumn{1}{c}{\textbf{39.73}}    & \multicolumn{1}{c}{\textbf{30.23}}  & \textbf{34.25}    \\ \hline
\end{tabular}%
}
\end{table*}

\subsection{Semantic label extraction}
The VLSP-2021 corpus includes both English and Vietnamese. Therefore, the extraction of the semantic label and data preprocessing for English and Vietnamese are performed separately. Then, they will be combined to get a complete semantic information corpus.




For Vietnamese, the process of semantic extraction is more complicated. The ten XLM-R\textsuperscript{1} models have been fine-tuned for the problem of SRL on the LORELEI corpus to predict semantic labels for the VLSP-2021 corpus on the Vietnamese domain. Next, the output data of 10 models will be combined. After that, we proceed to the data preprocessing stage, which includes the following steps:
\begin{itemize}
    \item Eliminate duplicate data points of output data.
    \item Eliminate data points whose label sequence contains two verb labels simultaneously.
    \item Combine sequences of semantic labels that belong to the same sentence.
\end{itemize}

For English, we use the XLM-R\textsuperscript{2} model fine-tuned on the ConLL12 corpus, which is part of Oliveira's work \cite{oliveira2021improving}. After predicting semantic labels for the VLSP-2021 corpus in the English domain, we perform the data preprocessing stage. The data preprocessing phase mainly checks and removes the data points the model cannot predict.


\subsection{Fine-tuning of SemBERT variant models for RTE}

\begin{table}[!ht]
\centering
\caption{Hyperparameters Are Set To Fine Tune The RTE problem.}
\label{table:Hyperparameter}
\resizebox{8.0cm}{!}{%
\begin{tabular}{|l||ccc|}
\hline
              & mBERT & XLM-R\textsubscript{base} & XLM-R\textsubscript{large} \\ \hline \hline
learning rate & $2e-5$  & $2e-5$                  & $1e-5$                   \\ \hline
weigth decay  & $0.01$  & $0.01$                  & $0.01$                   \\ \hline
batch size    & $12$    & $12$                    & $12$                     \\ \hline
max length   & $256$   & $256$                   & $256$                    \\ \hline
epoch         & $5$     & $5$                     & $5$                      \\ \hline
\end{tabular}%
}
\end{table}

The two vectors representing semantic and contextual representation will be combined to form a vector representing semantic cognitive context, \textit{h}. The vector \textit{h} will be passed directly through the classification class to generate the prediction probability corresponding to agree, disagree, and neutral labels. The training corpus VLSP-2021 will be divided into two sets for training and validation with a ratio of 8/2 with equal distribution of the number of labels and languages on these two sets. The cost function for this fine-tuning process is CrossEntropy; the AdamW optimization algorithm and the hyperparameters are set up as shown in Table \ref{table:Hyperparameter}.


\section{Result}

\begin{table*}[ht]
\centering
\caption{Predictive Results On The VLSP-2021 Test Corpus For The RTE Problem ($\%)$.}
\label{table:KetQuaTapTest}
\resizebox{11cm}{!}{%
\begin{tabular}{|l||cccc|}
\hline
\multicolumn{1}{|c||}{Model} & Accuracy & Precision & Recall & F\textsubscript{1}-score \\ \hline \hline
mBERT                         & 74.56    & 75.02     & 74.56  & 74.70    \\ \hline
XLM-R\textsubscript{base}                    & 81.14    & 81.17     & 81.16  & 81.16    \\ \hline
XLM-R\textsubscript{large}                   & 87.92    & 87.94     & 87.94  & 87.91    \\ \hline
mBERT + Semantic              & 76.07    & 76.22     & 76.07  & \textbf{76.12}    \\ \hline
XLM-R\textsubscript{base} + Semantic         & 81.34    & 81.38     & 81.36  & \textbf{81.25}    \\ \hline
XLM-R\textsubscript{large} + Semantic        & 88.91    & 88.91     & 88.93  & \textbf{88.89}    \\ \hline
\end{tabular}%
}
\end{table*}

Table \ref{table:KetQuaTapTest} shows the performance evaluation of the models on the VLSP-2021 test corpus. The results show that integrating semantic representation through the SRL problem with context representation of pre-training language models positively affects the results. Specifically, when fine-tuning the mBERT, XLM-R\textsubscript{base}, and XLM-R\textsubscript{large} models with a combination of semantic representations for the RTE problem, the results were 76.12\%, 81.25\%, and 88.89\%, respectively (F\textsubscript{1}-score). This result increased by 1.42\%, 0.09\%, and 0.98\%, respectively, when compared with the mBERT, XLM-R\textsubscript{base}, and XLM-R\textsubscript{large} models that did not incorporate semantic representation.

Table \ref{table:vi_en_results} shows the results of the VLSP-2021 test corpus for the RTE problem in each language. It can be seen from the table that the prediction results in Vietnamese of the mBERT, XLM-R\textsubscript{base}, and XLM-R\textsubscript{large} models, when combined with semantics, give better results than in the English domain reaching 76.22\%, 82.66\%, and 89.98\% (F\textsubscript{1}-score), respectively. Specifically, for each model, this result increased by 0.23\%, 2.99\%, and 2.26\%, respectively. On the other hand, considering the XLM-R\textsubscript{large} model when not incorporating semantic information, the results in the English domain are higher at 87.97\%. However, after combining the resulting semantic information, this model improved when it was higher than the original XLM-R\textsubscript{large} model. This result proves that Vietnamese is a semantically rich language, so enriching the context of Vietnamese texts improves the model's ability to understand natural language.

\begin{table*}[!ht]
\centering
\caption{Prediction Results On Test Corpus VLSP-2021 For The RTE Problem On Each Language ($\%$).}
\label{table:vi_en_results}
\resizebox{12cm}{!}{%
\begin{tabular}{|l||ccc||ccc|}
\hline
\multicolumn{1}{|c||}{\multirow{2}{*}{}}                          & \multicolumn{3}{c||}{Vietnamese}                                         & \multicolumn{3}{c|}{English}                                          \\ \cline{2-7} 
\multicolumn{1}{|c||}{}                                           & \multicolumn{1}{c}{Precision} & \multicolumn{1}{c}{Recall} & F\textsubscript{1}-score & \multicolumn{1}{c}{Precision} & \multicolumn{1}{c}{Recall} & F\textsubscript{1}-score \\ \hline \hline
mBERT                                                            & \multicolumn{1}{c}{75.66}     & \multicolumn{1}{c}{75.33}  & 75.35    & \multicolumn{1}{c}{74.47}     & \multicolumn{1}{c}{73.78}  & 73.97    \\ \hline
XLM-R\textsubscript{base}                                                       & \multicolumn{1}{c}{81.42}     & \multicolumn{1}{c}{81.43}  & 81.43    & \multicolumn{1}{c}{80.87}     & \multicolumn{1}{c}{80.87}  & 80.83    \\ \hline
XLM-R\textsubscript{large}                                                      & \multicolumn{1}{c}{87.89}     & \multicolumn{1}{c}{87.85}  & 87.86    & \multicolumn{1}{c}{87.98}     & \multicolumn{1}{c}{88.09}  & 87.97    \\ \hline \hline
\begin{tabular}[c]{@{}l@{}}mBERT + Semantic\end{tabular}       & \multicolumn{1}{c}{76.33}     & \multicolumn{1}{c}{76.27}  & 76.22    & \multicolumn{1}{c}{76.15}     & \multicolumn{1}{c}{75.90}  & 75.99    \\ \hline
\begin{tabular}[c]{@{}l@{}}XLM-R\textsubscript{base} + Semantic\end{tabular}  & \multicolumn{1}{c}{82.81}     & \multicolumn{1}{c}{82.72}  & 82.66    & \multicolumn{1}{c}{79.80}     & \multicolumn{1}{c}{79.90}  & 79.67    \\ \hline
\begin{tabular}[c]{@{}l@{}}XLM-R\textsubscript{large} + Semantic\end{tabular} & \multicolumn{1}{c}{90.04}     & \multicolumn{1}{c}{89.97}  & 89.98    & \multicolumn{1}{c}{87.68}     & \multicolumn{1}{c}{87.80}  & 87.72    \\ \hline
\end{tabular}%
}
\end{table*}

\subsection{Error analysis}
\begin{figure}[ht]
\centering
\includegraphics[scale=0.32]{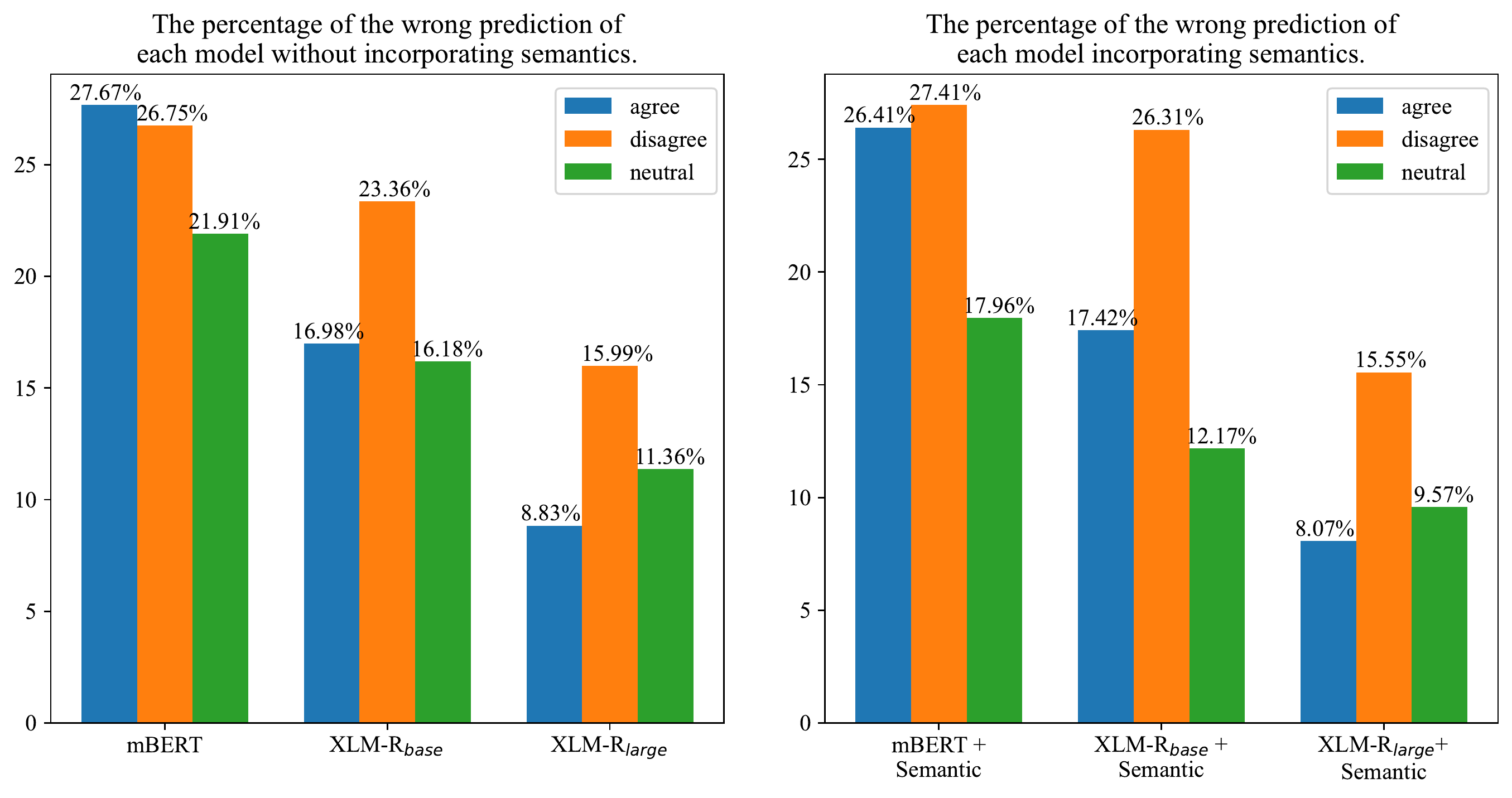}
\caption{The Percentage Of Wrong Predictions Of The Models On Each Label.}
\label{fig:WrongPredictionPercentage}
\end{figure}

Figure \ref{fig:WrongPredictionPercentage} depicts the wrong prediction trend of the mBERT, XLM-R\textsubscript{base}, and XLM-R\textsubscript{large} models without semantic matching and those with semantic matching on each label of the VLSP-2021. It can be observed that all three models, including semantically integrated mBERT, XLM-R\textsubscript{base}, and XLM-R\textsubscript{large}, for good prediction in the \textit{neutral} label. Specifically, the wrong prediction rate decreases by 3.95\%, 4.01\%, and 1.79\%, respectively, compared with models that do not incorporate semantics. However, the \textit{disagree} label error rates of the mBERT and XLM-R\textsubscript{base} models combined with semantics increased by 0.66\% and 2.95\%, respectively. For the \textit{agree} label, the error rate of the mBERT and XLM-R\textsubscript{large} models with combined semantics decreased by 1.26\% and 0.76\%, respectively. In all three semantic association model models, the XLM-R\textsubscript{large} model has improved the error rate of all three labels.

\subsection{The Influence Of Number Of Predicates}

\begin{table}[ht]
\centering
\caption{Experimental Results On The Influence Of The Number Of Predicates Of The Label Sequence When It Is Included In The Model ($\%$).}
\label{table:SoLuongViTu}
\resizebox{8.6cm}{!}{%
\begin{tabular}{|@{\hspace*{3mm}}l||*{5}{c}|}\hline
\multicolumn{1}{|@{}l||}{\backslashbox[1pt][l]{Model}{No of predicates}}
&\makebox[3em]{1}&\makebox[3em]{2}&\makebox[3em]{3}
&\makebox[3em]{4}&\makebox[3em]{5}\\\hline\hline
mBERT + Semantic & 74.83 & \textbf{76.12} & 75.10& 75.90 & 75.77 \\ \hline
XLM-R\textsubscript{base} + Semantic & 80.49 & 81.01& \textbf{81.25} & 80.12 & 80.51 \\ \hline
XLM-R\textsubscript{large} + Semantic & 87.83 & \textbf{88.89} & 88.27 & 88.21& 87.45 \\ \hline
\end{tabular}%
}
\end{table}

The predicate or verb is an essential component for the SRL problem; the predicate can determine the semantic role of the related components. Therefore, the number of predicates of the input text can affect model performance. Therefore, based on this experimental strategy, we also conduct experiments to conclude. Table \ref{table:SoLuongViTu} shows that with the predicate number of $2$, the mBERT and XLM-R-large model combined semantics give the best results. Meanwhile, the XLM-R-base model has the best performance with some predicates of $3$. This result can explain that in the natural language, it is scarce that the text contains many verbs at the same time. Therefore, the large number of predicates may be due to the confusion of the semantic role labeling models in the semantic extraction process.


\section{Conclusion}
We have presented the implementation process and experimental results for the RTE problem. The comparison results show that the mBERT, XLM-R\textsubscript{base}, and XLM-R\textsubscript{large} models that combine semantic representation have better performance when compared to primitive context representation models. Specifically, the performance of pre-training models with contextual and semantic representation has increased by about 1\%. However, while doing this experiment, we also had difficulties accessing the corpus and the limitation on the corpus size for the SRL problem, specifically the LORELEI corpus. Therefore, the fine-tuning of the XLM-R and RemBERT model for the SRL problem is not good. This work affects the extraction of semantic labels on the VLSP-2021 corpus in Vietnamese. As a result, the fine-tuning of SemBERT variant models for the RTE problem can also be affected.

Through the experimental results, it is possible that the critical role of shallow text representation through the SRL task in NLU. Especially in Vietnamese, there are currently not many works on corpus for this task. Going through the promising results based on the SRL problem of RTE can create a premise to develop this work in Vietnamese.

\section*{Acknowledgement}
This research is funded by Vietnam National University HoChiMinh City (VNU-HCM) under grant number DS2022-26-01.

\bibliographystyle{IEEEtran}
\bibliography{main.bib}

\end{document}